\documentclass[preprint,12pt]{elsarticle}

\usepackage{amsmath, amsfonts}
\usepackage{graphicx}
\usepackage{multirow}
\usepackage{booktabs}
\usepackage{siunitx}
\usepackage{url}
\usepackage{hyperref}
\usepackage{xcolor}

\journal{Atmospheric Environment} 

\begin{document}

\begin{frontmatter}

\title{Physics Informed Reconstruction of Four-Dimensional Atmospheric Wind Fields Using Multi-UAS Swarm Observations in a Synthetic Turbulent Environment}

\author[ame]{Abdullah Tasim}
\author[ame]{Wei Sun}

\address[ame]{School of Aerospace and Mechanical Engineering, University of Oklahoma, Norman, OK 73069, USA}

\begin{abstract}
Accurate reconstruction of atmospheric wind fields is critical for applications such as weather monitoring, hazard prediction, and wind energy assessment, yet conventional instruments leave spatio-temporal gaps within the lower atmospheric boundary layer. Unmanned aircraft systems provide flexible in situ measurements, but individual platforms sample wind only along their flight trajectories, making full wind-field recovery challenging.

This study presents a framework for reconstructing four-dimensional atmospheric wind fields using measurements obtained exclusively from a coordinated swarm of UAS. A synthetic turbulence environment and high-fidelity multirotor simulation are used to generate training and evaluation data. Local wind components are estimated from UAS dynamics using a bidirectional long short-term memory network (Bi-LSTM), and these estimates are assimilated into a physics-informed neural network (PINN) that reconstructs a continuous wind field as a function of space and time.

For local wind estimation, the bidirectional LSTM achieves root-mean-square errors (RMSE) of 0.064 and 0.062~m~s$^{-1}$ for the north and east components in low-wind conditions, 0.122 and 0.129~m~s$^{-1}$ under moderate-wind conditions, and 0.273 and 0.271~m~s$^{-1}$ in high-wind conditions. The vertical wind component exhibits higher error, with RMSE values of 0.029, 0.061, and 0.091~m~s$^{-1}$ across the same regimes, reflecting reduced observability from multirotor dynamics.

The physics-informed reconstruction successfully recovers the dominant spatial and temporal structure of the wind field up to 1000~m altitude, preserving mean flow direction and vertical shear while smoothing unresolved turbulent fluctuations. Under moderate wind conditions, the reconstructed mean wind field achieves overall RMSE between 0.118 and 0.154~m~s$^{-1}$ across the evaluated UAS configurations. The lowest reconstruction error of 0.118~m~s$^{-1}$ is obtained using a five-UAS configuration, while larger swarms do not yield monotonic improvement in accuracy. These results demonstrate that coordinated UAS measurements enable accurate and scalable four-dimensional wind-field reconstruction without relying on dedicated wind sensors or fixed infrastructure.

\end{abstract}

\begin{keyword}

UAS swarm \sep Local wind estimation (NED)\sep Long Short Term Memory (LSTM) \sep Physics-informed neural networks (PINN) \sep atmospheric boundary layer (ABL) \sep turbulence simulation
\end{keyword}

\end{frontmatter}

\section{Introduction}

Understanding how wind varies across space and time is key fundamental to a wide range of atmospheric and engineering applications, including numerical weather prediction, storm hazard assessment, and wind energy assessment. Regions that frequently experience strong winds are often the same areas most susceptible to high-impact events such as severe storms and tornadoes, making accurate characterization of wind fields important for both public safety and infrastructure planning. Although Doppler wind lidars, meteorological towers, and anemometers provide valuable wind measurements, their use is limited by deployment cost, operational constraints, and a rapid decline in coverage above a few hundred meters. As a result, large portions of the lower atmospheric boundary layer remain sparsely observed in both spatial and temporal dimensions \cite{william2021,barthelmie,Tasim}.

UAS have emerged as a complementary platform for atmospheric sensing because they provide high-resolution in situ observations and can be deployed flexibly at comparatively low cost \cite{Ivey,barbieri,Elston14}. However, measurements collected by a single UAS are confined to its flight trajectory, and local wind estimates derived from such paths do not reveal the broader three-dimensional structure of the surrounding flow. To support applications such as wind-energy resource assessment, turbine design, and characterization of boundary-layer dynamics, it is therefore necessary to reconstruct a complete four-dimensional wind field that evolves continuously in space and time.

Existing efforts toward wind-field reconstruction have largely followed two complementary directions. The first direction relies on supervised learning methods or reduced-order representations to infer flow structure from limited measurements. In the context of wind-energy applications, data-driven mapping models have been developed to estimate wind characteristics and to improve understanding of inflow variability and its influence on turbine power production \cite{3DWind,3DPriorwork,WindResourceAssesment,KIMwindTurbinepowerchange}. While these approaches can provide practical predictive capability, their performance is often sensitive to the representativeness of the training data, and they may not enforce physical consistency when measurements are sparse or operating conditions differ from those seen during training.

A second direction integrates physical constraints directly into the reconstruction process. PINN has been introduced as a mesh-free framework that blends sparse observations with governing equations by penalizing the residual of the underlying partial differential equations during training \cite{RAISSI}. This class of methods is particularly attractive for wind-field reconstruction because it regularizes the solution in under-sampled regions using physically meaningful constraints rather than relying purely on interpolation.

Several wind-focused studies have combined DWL measurements with physics-informed learning to reconstruct spatiotemporal flow fields. DWL systems are appealing because they enable remote sampling across planes or volumes and can provide repeated observations over time. However, DWL measurements are typically available as line-of-sight wind speeds, which represent projections of the wind vector onto the beam direction. Consequently, the inverse reconstruction problem can become ill-conditioned when only limited beam geometries are available. To address this challenge, physics-informed formulations have been proposed that assimilate sparse line-of-sight measurements while enforcing fluid-dynamic structure through equation-based constraints, enabling recovery of coherent wind features across space and time \cite{Lidarbased1,Lidarbased2,Lidarbased3andPOD}. These studies demonstrate that physics-based regularization can substantially improve reconstruction fidelity relative to purely data-driven fitting when measurements are sparse.

Despite these advances, DWL-centered physics-informed reconstructions remain constrained by practical sensing geometry and data type. Line-of-sight measurements do not directly provide horizontal wind components everywhere in the domain, and reconstruction accuracy depends strongly on scan strategy, sampling density, and the availability of multiple beams. In addition, DWL signal quality can degrade under conditions commonly encountered in the ABL, which further reduces the effective information content available for reconstruction \cite{Lidarbased1,Lidarbased3andPOD}.

These limitations motivate the use of UAS measurements as an alternative information source for physics-informed wind-field reconstruction. Unlike DWL, which primarily provides projected velocities, a UAS can sample the flow in situ along programmable three-dimensional trajectories and can supply wind estimates in component form after onboard estimation. From a physics-informed learning perspective, component-wise wind observations provide a more direct anchor for the reconstructed state variables, reducing ambiguity in the inverse problem and decreasing reliance on sensing geometry assumptions. Moreover, coordinated multi-UAS sampling enables spatial coverage to be distributed adaptively in both the horizontal and vertical directions, which is particularly beneficial for resolving boundary-layer shear, directional veer, and evolving turbulent structures that are difficult to capture using fixed or beam-limited sensors. As a result, physics-informed reconstruction driven by UAS-based measurements offers a practical pathway toward multi-dimensional wind-field reconstruction with improved flexibility in sampling design and higher information density per unit deployment \cite{RAISSI,Lidarbased1,Lidarbased3andPOD,3DWind}.

This work addresses these limitations by introducing a wind-field reconstruction framework that relies exclusively on measurements collected by a coordinated swarm of UAS and enables reconstruction of a fully four-dimensional wind field in space and time. In contrast to prior physics-informed wind reconstruction studies that assimilate fixed or remote-sensing observations and are typically restricted to horizontal flow structure, the proposed approach reconstructs wind as a function of $(x,y,z,t)$ using only mobile in situ measurements from UAS. The framework integrates a synthetic atmospheric turbulence environment, a high-fidelity multirotor dynamics simulator, a LSTM network for estimating local wind from UAS dynamics, and a PINN that fuses these estimates with weak physical constraints to recover the global wind field.

A key distinction of the proposed method is that each UAS infers local wind without requiring any dedicated wind sensor, relying solely on its intrinsic dynamical response to atmospheric disturbances. This eliminates dependence on specialized instrumentation and enables wind reconstruction in environments where conventional sensing infrastructure is impractical or unavailable. The results demonstrate that coordinated UAS measurements provide sufficient information to reconstruct a time-varying three-dimensional wind field up to 1000~m altitude, establishing a pathway toward autonomous, sensor-light, UAS-based four-dimensional atmospheric wind sensing.

\section{Synthetic Wind Model}
\label{sec:wind_model}

To reconstruct the atmospheric wind field from UAS-based local wind measurements requires a realistic yet computationally feasible representation of atmospheric turbulence. In this work, a synthetic wind environment is constructed by superimposing a vertically sheared mean wind profile with directional veer and a three-dimensional, zero-mean turbulent fluctuation field. The turbulent component is generated using a von K\'arm\'an spectral model, which is widely used in aircraft simulation and has been shown to reproduce the statistics of continuous atmospheric turbulence at low altitudes relevant to small UAS operations \cite{WangFrost1980,Cole2019SpatioTemporalWind}.

\subsection{Atmospheric turbulence representation}

The instantaneous wind velocity vector $\boldsymbol{V}(\boldsymbol{x},t)$ at position $\boldsymbol{x}=(x,y,z)$ and time $t$ is decomposed into a mean and a fluctuating component,
\begin{equation}
  \boldsymbol{V}(\boldsymbol{x},t)
  = \overline{\boldsymbol{V}}(z,t) + \boldsymbol{v}'(\boldsymbol{x},t),
\end{equation}
where $\overline{\boldsymbol{V}}(z,t)$ represents the horizontally homogeneous mean wind, assumed to depend primarily on height $z$ and slowly on time, and $\boldsymbol{v}'$ denotes the turbulent fluctuations. The statistical properties of $\boldsymbol{v}'$ are specified through power spectral densities (PSDs) and turbulence length scales, following standard spectral turbulence models for aviation applications \cite{vonKarman1948, Cole2019SpatioTemporalWind}.

Under the usual assumptions of homogeneous, stationary turbulence and Taylor's frozen-flow hypothesis \cite{TaylorsHypothesisandHighFrequencyTurbulenceSpectra}, the longitudinal, lateral, and vertical turbulent velocity components $(u',v',w')$ can be treated as stochastic processes with prescribed PSDs as functions of spatial frequency. For low-altitude flight, the von K\'arm\'an model defined in military specifications such as MIL-F-8785C provides height-dependent turbulence intensities and length scales that have been validated for both conventional aircraft and small UAS \cite{WangFrost1980,patel2008,watkins2006}.%

\subsection{Von K\'arm\'an spectral model}

The von K\'arm\'an model characterizes turbulence through PSDs that are consistent with Kolmogorov's $-5/3$ inertial-subrange scaling, providing a better match to measured atmospheric spectra than simpler rational approximations such as the Dryden model \cite{vonKarman1948,Agren2020}.%
For an aircraft (or UAS) flying through a frozen turbulence field with longitudinal speed $U_0$, the PSDs of the turbulence components as functions of circular frequency $\omega$ can be written in the low-altitude von K\'arm\'an form \cite{Cole2019SpatioTemporalWind}:
\begin{align}
  \Phi_u(\omega) &=
  \frac{2 \sigma_u^2 L_u}{\pi U_0}
  \left[1 + \left( 1.339 \frac{L_u \omega}{U_0} \right)^2 \right]^{-5/6}, \\
  \Phi_v(\omega) &=
  \frac{\sigma_v^2 L_v}{\pi U_0}
  \frac{1 + \frac{8}{3}\left( 1.339 \frac{L_v \omega}{U_0} \right)^2}
       {\left[1 + \left( 1.339 \frac{L_v \omega}{U_0} \right)^2 \right]^{11/6}}, \\
  \Phi_w(\omega) &=
  \frac{\sigma_w^2 L_w}{\pi U_0}
  \frac{1 + \frac{8}{3}\left( 1.339 \frac{L_w \omega}{U_0} \right)^2}
       {\left[1 + \left( 1.339 \frac{L_w \omega}{U_0} \right)^2 \right]^{11/6}}.
\end{align}
Here, $L_u$, $L_v$, and $L_w$ are the longitudinal, lateral, and vertical turbulence length scales, and $\sigma_u$, $\sigma_v$, and $\sigma_w$ are the corresponding turbulence intensities (standard deviations of $u'$, $v'$, and $w'$). In the low-altitude formulation, these quantities are expressed as functions of height $z$ and a reference wind speed $u_{20}$ at 20~ft (6~m), ensuring consistency with boundary-layer turbulence measurements \cite{Cole2019SpatioTemporalWind}.%

In the present work, an isotropic three-dimensional von K\'arm\'an velocity field is generated in a finite domain using a spectral method. Following the approach of Cole and Wickenheiser \cite{Cole2019SpatioTemporalWind}, a set of random Fourier modes is drawn from the von K\'arm\'an energy spectrum and superposed to obtain a zero-mean, divergence-free turbulent field $(u'(\boldsymbol{x}),v'(\boldsymbol{x}),w'(\boldsymbol{x}))$. This isotropic field is then rescaled to match prescribed turbulence intensity levels, and subsequently interpolated in space and time along each UAS trajectory in the simulation domain. 

Concretely, a cubic domain of size $L_x=L_y=210$~m and $L_z=1010$~m is discretized onto a $64\times64\times128$ grid. A von K\'arm\'an spectrum $E(k)$ is evaluated as a function of wavenumber $k$, and a finite number of modes is sampled to construct the turbulent field via the isotropic generator in \texttt{TurboGenPY}, which provides three-dimensional realizations of von K\'arm\'an turbulence \cite{Saad2016,Saad2016v2,Richards2018}. The resulting real-valued fields are normalized by their initial standard deviation $\sigma_0$ and scaled to the desired turbulence intensity at each height, as described in this section.%

\subsection{Mean wind, shear, and veer}

The turbulent fluctuations are superimposed on a mean wind profile that captures vertical shear, directional veer, and a low-frequency bias representing slow gust-like modulation. The mean horizontal wind speed at height $z$ is prescribed using a power-law profile,
\begin{equation}
  U(z) = U_{\mathrm{ref}}
  \left( \frac{\max(z,z_{\mathrm{ref}})}{z_{\mathrm{ref}}} \right)^{\alpha},
\end{equation}
where $U_{\mathrm{ref}}$ is the reference wind speed at height $z_{\mathrm{ref}}=10$~m and $\alpha$ is a shear exponent that sets the strength of vertical variation. A height-dependent wind direction $\gamma(z)$ is used to represent directional veer,
\begin{equation}
  \gamma(z) = \gamma_0 + \kappa_{\mathrm{veer}} \frac{z}{100},
\end{equation}
where $\gamma_0$ is the reference wind direction at the surface and $\kappa_{\mathrm{veer}}$ specifies the veer rate in degrees per 100~m.

To introduce slow temporal variability in the background wind, a sinusoidal bias is added,
\begin{equation}
  U_{\mathrm{eff}}(z,t) = U(z) + \Delta U \sin\left( \frac{2\pi t}{T_{\mathrm{gust}}} \right),
\end{equation}
with amplitude $\Delta U$ and period $T_{\mathrm{gust}}$. This term represents mesoscale fluctuations on time scales longer than the turbulent eddies, providing a simple parameterization of gust cycles without explicitly modeling discrete gust structures \cite{Cole2019SpatioTemporalWind,Richards2018}.

The effective mean wind vector is then written in horizontal components aligned with the North–East–Down (NED) axes,
\begin{align}
  U_N(z,t) &= -U_{\mathrm{eff}}(z,t)\cos\gamma(z), \\
  U_E(z,t) &= -U_{\mathrm{eff}}(z,t)\sin\gamma(z),
\end{align}
and combined with the turbulent fluctuations to form the total wind,
\begin{equation}
  \boldsymbol{V}(\boldsymbol{x},t)
  = \begin{bmatrix} U_N(z,t) \\ U_E(z,t) \\ 0 \end{bmatrix}
  + \begin{bmatrix} u'(\boldsymbol{x}-\boldsymbol{U} t) \\ v'(\boldsymbol{x}-\boldsymbol{U} t) \\ w'(\boldsymbol{x}-\boldsymbol{U} t) \end{bmatrix},
\end{equation}
where the argument shift $\boldsymbol{x}-\boldsymbol{U}t$ implements Taylor's frozen-field hypothesis by advecting the turbulent field past the UAS at the mean wind speed. Spatial interpolation is used to evaluate the turbulent components at arbitrary positions along the UAS trajectories.%

\subsection{Rationale for model selection}

Several classes of wind models were considered for constructing the synthetic environment, including the Dryden turbulence model, large-eddy simulations (LES) of the atmospheric boundary layer, and more recent fractional or hybrid spectral models \cite{Agren2020,Nithya2024,Hajjem2022}.
The von K\'arm\'an model was selected for three main reasons:

\begin{enumerate}
  \item \textbf{Spectral realism at low altitude.} Compared to the Dryden model, which uses rational PSDs with a steeper high-frequency roll-off, the von K\'arm\'an spectra reproduce the $-5/3$ slope of the inertial subrange and have been shown to match measured low-altitude turbulence more closely \cite{WangFrost1980, Cole2019SpatioTemporalWind}. This is especially important when evaluating UAS controllers and local wind estimators that are sensitive to high-frequency fluctuations.%

  \item \textbf{Computational efficiency versus CFD.} LES and related computational fluid dynamics (CFD) approaches can resolve detailed urban or complex-terrain flows \cite{Nithya2024}, but they are computationally intensive and not practical for generating the thousands of wind realizations required for training and testing data-driven models. The von K\'arm\'an spectral approach provides a middle ground: it preserves realistic frequency content while remaining inexpensive enough to run on a personal workstation.%

  \item \textbf{Compatibility with 4D reconstruction.} The isotropic von K\'arm\'an field can be generated in three dimensions and combined with analytic mean profiles for shear, veer, and temporal modulation. This flexibility is essential for the present study, where the goal is to reconstruct a four-dimensional wind field (space and time) from UAS-based local wind estimates using a PINN. The spectral formulation also makes it straightforward to control turbulence intensity, integral length scales, and mean wind parameters across different test cases. 
\end{enumerate}

More advanced models, such as fractional spectral formulations that refine von K\'arm\'an fits at very low altitudes \cite{Hajjem2022}, are promising for future work but introduce additional parameters and calibration requirements that are beyond the scope of this initial methodology study.%
For the current objective---developing and evaluating a complete pipeline from UAS dynamics to local wind estimation and four-dimensional wind-field reconstruction---the von K\'arm\'an model offers an appropriate balance between physical realism, controllability, and computational cost.

\section{UAS Dynamics Model, Reference Trajectory, and Control Architecture}
\label{sec:uas_dynamics}

This section describes the multirotor dynamics, reference trajectories, and control laws used to simulate the motion of each UAS in the swarm. The goal of the simulation is not to reproduce every aerodynamic feature of a real quadrotor, but to generate a physically consistent, high-fidelity platform whose responses to wind disturbances can be used to train a data-driven local wind estimator and to evaluate the wind-field reconstruction. 

\subsection{Rigid-Body Dynamics Model}

Each UAS is modeled as a rigid body with six degrees of freedom (6-DOF). The state of the vehicle consists of its position $\mathbf{p}=[x,y,z]^\top$ in the inertial North–East–Down (NED) frame, velocity $\mathbf{v}=[v_x,v_y,v_z]^\top$, attitude $\boldsymbol{\eta}=[\phi,\theta,\psi]^\top$, and angular velocity $\boldsymbol{\omega}=[p,q,r]^\top$ expressed in the body frame.

The translational dynamics follow directly from Newton’s second law:
\begin{equation}
m\dot{\mathbf{v}}
=
T_{\mathrm{tot}}\, \mathbf{z}_b + \mathbf{F}_{\mathrm{drag}} -
\begin{bmatrix}0 \\ 0 \\ mg\end{bmatrix},
\label{eq:trans_dyn}
\end{equation}
where $m=2.59~\mathrm{kg}$ is the vehicle mass, $T_{\mathrm{tot}}$ is the total thrust generated by the rotors, $\mathbf{z}_b$ is the third column of the rotation matrix $R(\phi,\theta,\psi)$ mapping body-frame thrust into the inertial frame, and $\mathbf{F}_{\mathrm{drag}}$ is the aerodynamic drag force described later in the section.

The rotational dynamics are given by
\begin{equation}
J\dot{\boldsymbol{\omega}}
=
\boldsymbol{\tau}
-
\boldsymbol{\omega} \times (J\boldsymbol{\omega}),
\label{eq:rot_dyn}
\end{equation}
where $J=\mathrm{diag}(0.078,\, 0.082,\, 0.14)$~kg·m$^2$ is the inertia matrix and $\boldsymbol{\tau}=[\tau_x,\tau_y,\tau_z]^\top$ is the control torque vector.

The Euler-angle kinematics relate the body angular rates $\boldsymbol{\omega}=[p,q,r]^\top$ to the attitude rates $\dot{\boldsymbol{\eta}}=[\dot{\phi},\dot{\theta},\dot{\psi}]^\top$ through
\begin{equation}
\dot{\boldsymbol{\eta}} = E^{-1}(\phi,\theta)\,\boldsymbol{\omega},
\end{equation}
where the Euler-rate matrix for the $3$--$2$--$1$ (yaw--pitch--roll) convention is
\begin{equation}
E^{-1}(\phi,\theta)
=
\begin{bmatrix}
1 & \sin\phi \tan\theta & \cos\phi \tan\theta \\
0 & \cos\phi            & -\sin\phi \\
0 & \dfrac{\sin\phi}{\cos\theta} & \dfrac{\cos\phi}{\cos\theta}
\end{bmatrix}.
\end{equation}

The attitude rotation matrix $R_{zyx}(\phi,\theta,\psi)$ mapping body-frame vectors
to the inertial North–East–Down frame is implemented using the standard 3--2--1
(yaw--pitch--roll) Euler sequence,
\begin{equation}
R_{zyx}(\phi,\theta,\psi)
=
\begin{bmatrix}
c_\psi c_\theta &
c_\psi s_\theta s_\phi - s_\psi c_\phi &
c_\psi s_\theta c_\phi + s_\psi s_\phi \\
s_\psi c_\theta &
s_\psi s_\theta s_\phi + c_\psi c_\phi &
s_\psi s_\theta c_\phi - c_\psi s_\phi \\
- s_\theta &
c_\theta s_\phi &
c_\theta c_\phi
\end{bmatrix},
\end{equation}
where $c_\cdot=\cos(\cdot)$ and $s_\cdot=\sin(\cdot)$.

\subsection{Rotor Thrust and Torque Generation}

Each of the four rotors produces thrust proportional to the square of its angular velocity:
\begin{equation}
f_i = k_T \omega_i^2, \qquad k_T = 1.1\times 10^{-5}~\mathrm{N\,s^2\,rad^{-2}}.
\end{equation}

The total thrust is
\begin{equation}
T_{\mathrm{tot}} = \sum_{i=1}^4 f_i.
\end{equation}

The body-frame torques produced by the differential thrusts are
\begin{align}
\tau_x &= L (f_4 - f_2),\\
\tau_y &= L (f_1 - f_3),\\
\tau_z &= \frac{k_M}{k_T} \sum_{i=1}^4 s_{yaw,i} f_i,
\end{align}
with arm length $L=0.25~\mathrm{m}$, torque coefficient $k_M = 0.055k_T$, and rotor spin directions
\[
s_{yaw} = [+1,-1,+1,-1].
\]

These expressions correspond exactly to the thrust–torque allocation implemented in the simulator.

\subsection{Motor Dynamics}

The simulation uses a first-order motor model in which each rotor speed
$\omega_i$ follows a commanded value $\omega_{i,\mathrm{cmd}}$ with a finite
response rate:
\begin{equation}
\dot{\omega}_i =
\frac{\omega_{i,\mathrm{cmd}} - \omega_i}{\tau_{\mathrm{motor}}},
\qquad
\tau_{\mathrm{motor}} = 0.10~\mathrm{s}.
\label{eq:motor_lag}
\end{equation}

The commanded rotor speeds are obtained directly from the thrust and torque
commands produced by the cascaded controller. The controller outputs the desired
total thrust $T_{\mathrm{cmd}}$ and body torques
$\boldsymbol{\tau}_{\mathrm{cmd}}=[\tau_x,\tau_y,\tau_z]^\top$. These four
quantities are converted into per-rotor thrusts by inverting the standard
allocation matrix $A$:
\begin{equation}
\mathbf{f}_{\mathrm{cmd}} = 
A^{-1}
\begin{bmatrix}
T_{\mathrm{cmd}} \\ \tau_x \\ \tau_y \\ \tau_z
\end{bmatrix},
\qquad
f_{i,\mathrm{cmd}}=\max(f_{i,\mathrm{cmd}},\,0).
\end{equation}

Each rotor thrust command is then mapped to a commanded rotor speed using
\begin{equation}
\omega_{i,\mathrm{cmd}} = 
\sqrt{\frac{f_{i,\mathrm{cmd}}}{k_T}},
\end{equation}
with $k_T=1.1\times10^{-5}~\mathrm{N\,s^2\,rad^{-2}}$. Finally, actuator
limits are enforced by saturating
\[
\omega_{\min}=0,\qquad \omega_{\max}=1200~\mathrm{rad/s}.
\]

This formulation ensures that rotor thrust is consistent with the multirotor
dynamics while capturing the finite acceleration capability of real motors.

\subsection{Aerodynamic Drag Model}

The only aerodynamic effect included in the dynamics is quadratic drag acting on the air-relative velocity:
\begin{equation}
\mathbf{F}_{\mathrm{drag}} =
- \tfrac{1}{2}\rho
\begin{bmatrix}
C_{d,xy} \\ C_{d,xy} \\ C_{d,z}
\end{bmatrix}
\!\cdot\!
|\mathbf{V}_{\mathrm{air}}| \, \mathbf{V}_{\mathrm{air}},
\end{equation}
where $\rho=1.225~\mathrm{kg/m^3}$, $C_{d,xy}=0.072$, $C_{d,z}=0.10$, and
\[
\mathbf{V}_{\mathrm{air}} = [v_x,v_y,v_z]^\top - [V_N,V_E,-V_D]^\top,
\]
with $(V_N,V_E,V_D)$ obtained from the synthetic wind field. No blade-flapping model, induced-velocity correction, or aerodynamic tilt effects are included, ensuring that the simulation matches the implemented code precisely.

\subsection{Reference Trajectory for Vertical Profiling}

All UAS follow the same prescribed vertical trajectory designed to emulate a typical atmospheric profiling mission. The reference altitude is given by
\begin{equation}
z_{\mathrm{ref}}(t) =
\begin{cases}
0, & t < 10~\mathrm{s},\\[4pt]
\displaystyle\frac{t - 10}{340}\,1000, & 10 \le t < 350,\\[6pt]
1000, & t \ge 350,
\end{cases}
\end{equation}
corresponding to a constant climb rate of approximately $2.94~\mathrm{m/s}$.

Each UAS maintains a constant horizontal reference $(x_{\mathrm{ref}}, y_{\mathrm{ref}})$ determined by its position in the swarm formation (e.g., a $3\times3$ grid for nine UAS). Because all vehicles track identical vertical profiles, variations in their measured wind responses can be attributed solely to spatial variability in the synthetic wind field, which is essential for training the LSTM-based estimator.

\section{Control System Design}

\begin{figure}[t]
    \centering
    \includegraphics[width=0.95\linewidth]{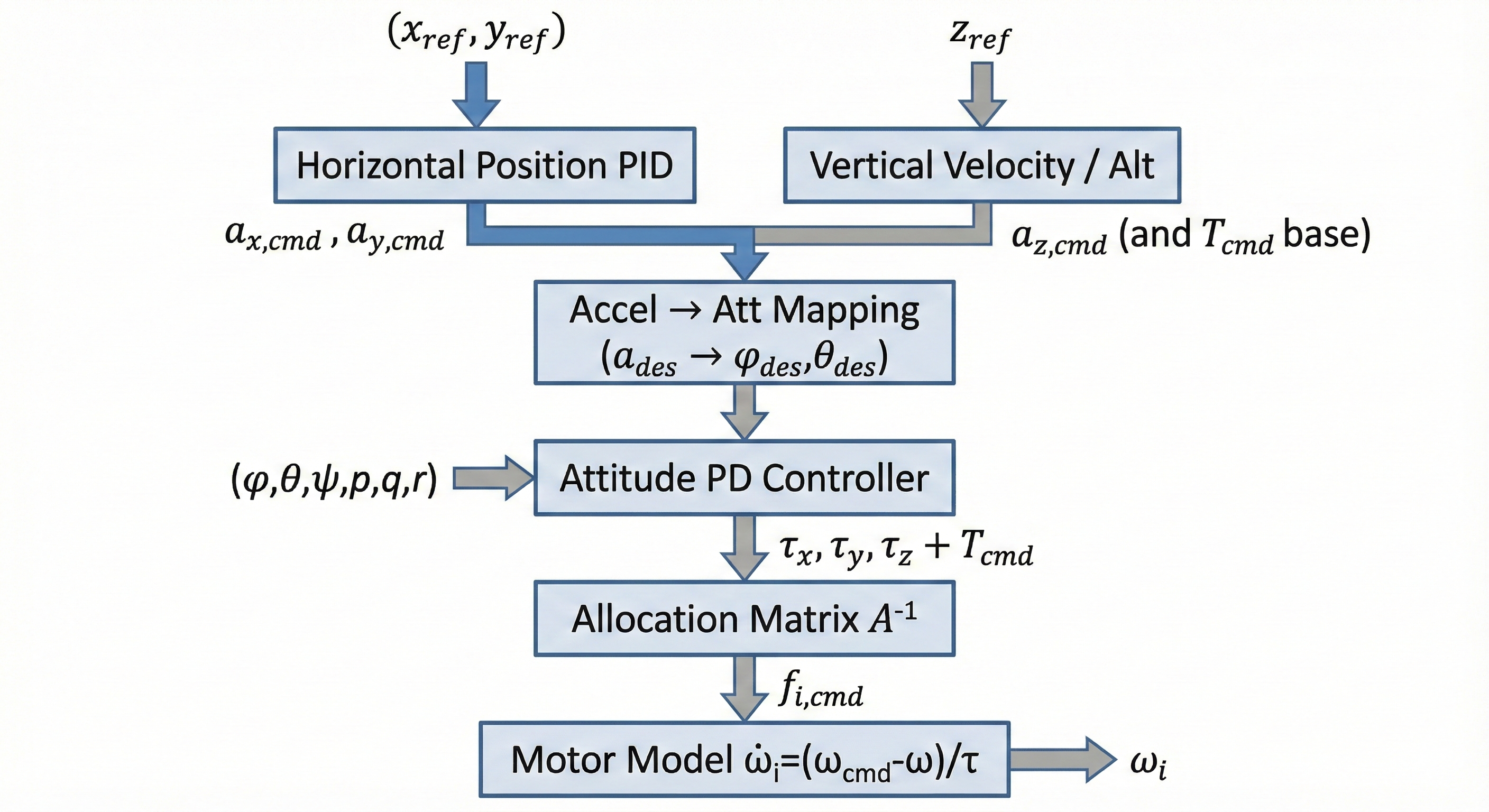}
    \caption{Cascaded control architecture used in the quadrotor simulation. 
    The controller consists of horizontal position regulation, altitude control, acceleration–to–attitude mapping, attitude PD 
    stabilization, rotor-force allocation, and motor dynamics.}
    \label{fig:controller}
\end{figure}

The quadrotor is stabilized using a cascaded control architecture that consists of an outer-loop position controller, an intermediate acceleration-to-attitude mapping, and an inner-loop attitude controller. The position loop regulates horizontal and vertical tracking errors and generates desired inertial accelerations $(a_{x,\mathrm{cmd}},a_{y,\mathrm{cmd}},a_{z,\mathrm{cmd}})$, with the vertical channel implemented as a velocity-limited altitude controller. These commanded accelerations are combined with gravity to form a desired specific force vector, whose direction defines the desired roll and pitch angles while yaw held at zero. The inner-loop attitude PD controller tracks these desired angles and produces the required body torques $(\tau_x,\tau_y,\tau_z)$ together with the collective thrust $T_{\mathrm{cmd}}$. A standard allocation matrix then converts the commanded thrust and torques into per-rotor thrusts, which are subsequently mapped to commanded rotor speeds. Each motor follows its commanded speed through a first-order response model, ensuring physically plausible actuator behavior. Figure \ref{fig:controller} shows the architecture of the cascaded control used for simulation.

\subsection{Position Control in the Horizontal Plane}

Horizontal motion is regulated using a PID controller acting on position errors:
\begin{equation}
e_x = x_{\mathrm{ref}} - x, \qquad
e_y = y_{\mathrm{ref}} - y.
\end{equation}

The raw commanded accelerations are
\begin{align}
a_{x,\mathrm{pid}} &= K_{p,x} e_x + K_{i,x} \int e_x dt + K_{d,x}(-v_x),\\
a_{y,\mathrm{pid}} &= K_{p,y} e_y + K_{i,y} \int e_y dt + K_{d,y}(-v_y),
\end{align}
with gains $(K_{p,x},K_{p,y})=(1.3,1.3)$, $(K_{i,x},K_{i,y})=(0.15,0.08)$, and $(K_{d,x},K_{d,y})=(1.4,1.4)$.

The total lateral acceleration magnitude is limited:
\begin{equation}
\sqrt{a_{x,\mathrm{pid}}^2 + a_{y,\mathrm{pid}}^2}
\le a_{xy,\max},
\qquad
a_{xy,\max} = 0.95\,g\,\tan(35^\circ).
\end{equation}

If saturation occurs, the integrators are back-calculated to prevent windup:
\begin{equation}
\int e_x dt \leftarrow
\int e_x dt -
K_{i,\mathrm{back,xy}}
\frac{a_{x,\mathrm{pid}} - a_{x,\mathrm{cmd}}}{K_{i,x}},
\end{equation}
with $K_{i,\mathrm{back,xy}}=0.5$. The same logic applies to the $y$-axis.

\subsection{Vertical Velocity Control}

To generate a vertical velocity target, the altitude error
\(
e_z = z_{\mathrm{ref}} - z
\)
is first scaled by a proportional gain \(k_v\). Because the climb and descent
rates are limited for both realism and controller stability, the resulting
velocity command is saturated to remain within a prescribed bound
\(V_{z,\max}\), i.e.,
\begin{equation}
v_{z,\mathrm{ref}} =
\operatorname{sat}\!\left(
k_v\, e_z,\;
[-V_{z,\max},\, V_{z,\max}]
\right),
\end{equation}
where \(k_v=1.1\) and \(V_{z,\max}=4~\mathrm{m/s}\).

The commanded vertical acceleration is
\begin{equation}
a_{z,\mathrm{temp}} = 
K_{pv}(v_{z,\mathrm{ref}} - v_z)
+ K_{i,v}\!\int e_z dt
- K_{dv} v_z,
\end{equation}
with $(K_{pv},K_{dv},K_{iv}) = (5.0,\,0.20,\,0.04)$.

Acceleration is clipped:
\[
|a_{z,\mathrm{temp}}| \le a_{z,\max}, \qquad a_{z,\max}=5~\mathrm{m/s^2},
\]
with anti-windup gain $K_{i,\mathrm{back,z}}=0.3$.

\subsection{Attitude Control and Thrust Direction}

The commanded acceleration vector is
\[
\mathbf{a}_{\mathrm{des}}
=
[a_{x,\mathrm{cmd}},\, a_{y,\mathrm{cmd}},\, g + a_{z,\mathrm{cmd}}]^\top.
\]
Normalizing this vector defines the desired body $z$-axis:
\[
\mathbf{z}_b^{\mathrm{des}} =
\frac{\mathbf{a}_{\mathrm{des}}}{\|\mathbf{a}_{\mathrm{des}}\|}.
\]

A desired roll and pitch are extracted from the desired orientation matrix $R_{\mathrm{des}}$, while yaw is held at zero.

The attitude controller is a PD law:
\begin{equation}
\boldsymbol{\tau}_{\mathrm{cmd}}
=
K_{p,\mathrm{att}}\,(\boldsymbol{\eta}_{\mathrm{des}}-\boldsymbol{\eta})
-
K_{d,\mathrm{att}}\,\boldsymbol{\omega},
\end{equation}
with
\[
K_{p,\mathrm{att}} = \mathrm{diag}(1.3,1.3,1.2),\quad
K_{d,\mathrm{att}} = \mathrm{diag}(0.9,0.9,0.7),
\]
and torque limits $\tau_{\max}=[4,4,3]~\mathrm{N\,m}$.

The commanded thrust is
\[
T_{\mathrm{cmd}}
=
\frac{m(g + a_{z,\mathrm{cmd}})}{\max(0.2,\,z_{b,z})}.
\]

\subsection{Rotor Allocation Matrix}

The thrust and torques are converted to rotor forces using the inverse of the allocation matrix
\[
A =
\begin{bmatrix}
1 & 1 & 1 & 1 \\
0 & -L & 0 & L \\
L & 0 & -L & 0 \\
(k_M/k_T)s_{yaw,1} & \dots & (k_M/k_T)s_{yaw,4}
\end{bmatrix},
\qquad
\mathbf{f}_{\mathrm{cmd}} = A^{-1}
\begin{bmatrix}
T_{\mathrm{cmd}} \\ \tau_x \\ \tau_y \\ \tau_z
\end{bmatrix}.
\]

Negative rotor forces are clipped to zero, then mapped to commanded rotor speeds:
\[
\omega_{\mathrm{cmd},i}
=
\sqrt{f_{\mathrm{cmd},i}/k_T}.
\]

\section{Local Wind Estimation from UAS Dynamics}
\label{lstm}

Accurate local wind estimation at each UAS is a critical component of the reconstruction of the wind field. As the instantaneous relation between UAS dynamics and wind in ABL is nonlinear and history-dependent, 
a data-driven recurrent model is adopted to infer the disturbance vector from onboard measurements. 
This section describes the feature construction, training dataset generation, and the Long Short-Term Memory (LSTM) network used to estimate the local wind components in the north–east–down (NED) frame.

\subsection{Feature Selection and Construction}

Each UAS records kinematic, dynamic, and controller-level quantities at a sampling frequency of 10~Hz. 
The feature vector at time $t$, denoted $\mathbf{x}(t)$, is constructed as
\begin{equation}
    \mathbf{x}(t) =
    \big[
        V_X,\,V_Y,\,V_Z,\,
        A_X,\,A_Y,\,A_Z,\,
        T_{\mathrm{cmd}},\,
        \phi,\,\theta,\,
        dX,\,dY
    \big]_t,
\end{equation}
where $(V_X,V_Y,V_Z)$ are inertial-frame velocities, $(A_X,A_Y,A_Z)$ are accelerations which could be obtained from the onboard IMU, $T_{\mathrm{cmd}}$ is the total commanded thrust, 
$\phi$ and $\theta$ are the roll and pitch angles, and $(dX,dY)$ represent horizontal tracking errors defined by
\begin{equation}
    dX = X - X_{\mathrm{ref}}, 
    \qquad
    dY = Y - Y_{\mathrm{ref}}.
\end{equation}
These terms embed information about controller effort and unmodeled disturbances, both of which correlate strongly with wind in the ABL.

The training labels consist of the filtered ground-truth wind components at the UAS locations obtained from the simulation stated in the earlier section \ref{sec:wind_model},
\begin{equation}
    \mathbf{w}(t) = 
    \big[
        U_N(t),\,
        U_E(t),\,
        U_D(t)
    \big],
\end{equation}
where a centered median filter of width 21 samples is applied to remove small-scale numerical noise and produce a smoothly varying target signal.  
This filtering step preserves the underlying mean and turbulence structure while stabilizing the learning process.

\subsection{Training Dataset Generation Across Wind Regimes}

\begin{figure}
    \centering
    \includegraphics[width=1.0\linewidth]{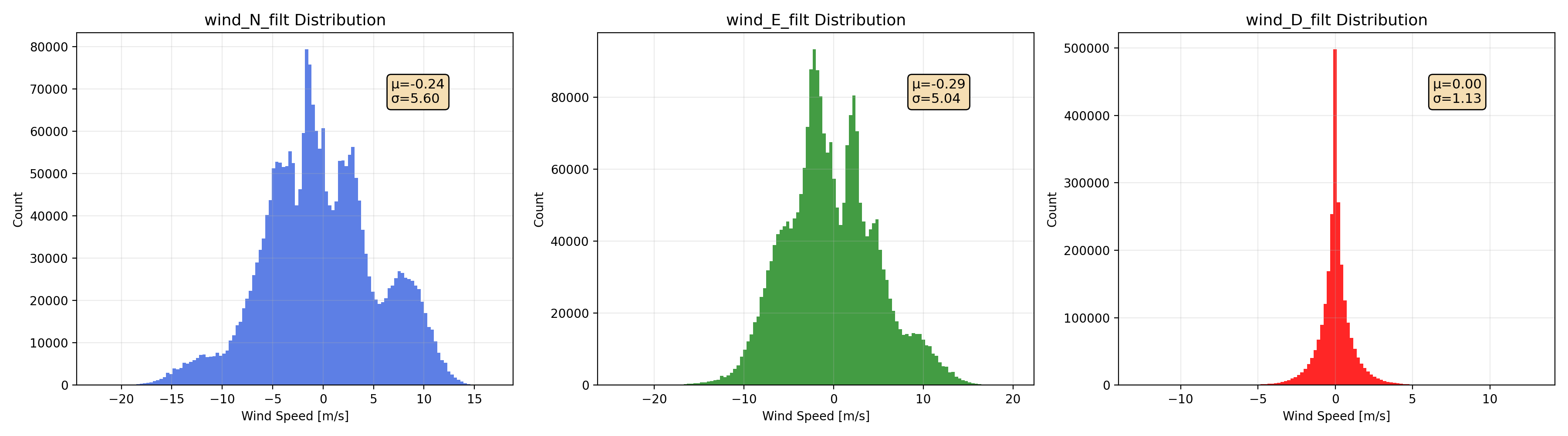}
    \caption{Histogram of the synthetic north, east, and down wind components extracted from all training flights. These distributions characterize the range of wind conditions used to generate the sequential training dataset.}
    \label{fig:dataset_generation}
\end{figure}

The LSTM is trained using a suite of synthetic flights generated under a range of atmospheric boundary-layer conditions. 
For each wind case, the simulator produces time-resolved trajectories for all UAS in the swarm, including the corresponding NED wind along their paths.  
Wind variability is introduced through changes in mean speed $U_{\mathrm{mean}}$, shear exponent $\alpha$, direction $\gamma$, vertical veer, and turbulence intensity $I_u$.
The resulting dataset contains more than $10^5$ sequential samples spanning 75 distinct wind realizations and 9 UAS per flight. However, the training was done in such a way that the model could be applied for any number of UAS configurations.Figure~\ref{fig:dataset_generation} shows the histogram of the north, east, and down wind components used for training the LSTM model. These distributions are obtained from all synthetic flights generated by the turbulence-driven simulation environment.

For each flight, sequences are extracted using a sliding window of length $L=40$ samples, corresponding to a 4~s temporal context:
\begin{equation}
    \mathcal{X}_i = \big\{\mathbf{x}(t_i), \mathbf{x}(t_i+1), \ldots, \mathbf{x}(t_i + L - 1)\big\},
\end{equation}
with the corresponding target wind vector
\begin{equation}
    \mathbf{y}_i = \mathbf{w}(t_i + L).
\end{equation}
All features are normalized using a standard z-score transformation:
\begin{equation}
    \tilde{\mathbf{x}} = \frac{\mathbf{x} - \mu_X}{\sigma_X},
    \qquad
    \tilde{\mathbf{y}} = \frac{\mathbf{y} - \mu_y}{\sigma_y}.
\end{equation}
Normalization is performed across the full multi-flight dataset to ensure consistent scaling across wind regimes.

\subsection{LSTM Network Architecture}

A bidirectional LSTM model is employed to map sequential feature histories to the instantaneous wind vector. 
The network contains two stacked LSTM layers with hidden dimension $h=256$, followed by a fully connected regression head:
\begin{align}
    \mathbf{h}_1 &= \mathrm{BiLSTM}_1(\tilde{\mathcal{X}}_i),\\[3pt]
    \mathbf{h}_2 &= \mathrm{BiLSTM}_2(\mathbf{h}_1),\\[3pt]
    \hat{\mathbf{y}}_i &= 
        \mathbf{W}_2 \,\sigma\!\left(
            \mathbf{W}_1 \mathbf{h}_2^{(L)} + \mathbf{b}_1
        \right) + \mathbf{b}_2,
\end{align}
where $\mathbf{h}_2^{(L)}$ is the final hidden state, and $\sigma(\cdot)$ denotes the ReLU activation.
Dropout layers with probability 0.2 are inserted between linear blocks to improve generalization.

Training minimizes a weighted mean-squared error,
\begin{equation}
    \mathcal{L} = 
    w_N \left(\hat{U}_N - U_N\right)^2 +
    w_E \left(\hat{U}_E - U_E\right)^2 +
    w_D \left(\hat{U}_D - U_D\right)^2,
\end{equation}

\begin{figure}
    \centering
    \includegraphics[width=1.0\linewidth]{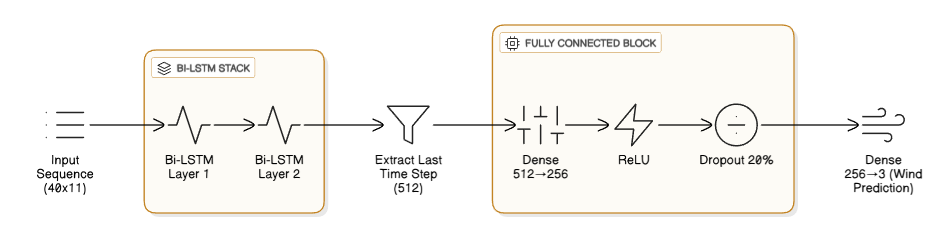}
    \caption{Architecture of the bidirectional LSTM model used for local wind estimation from UAS dynamics.}
    \label{Fig:Bilstm}
\end{figure}

where $(w_N,w_E,w_D) = (1.0,\,1.0,\,0.3)$ reduce the influence of vertical wind noise.
The Adam optimizer with an adaptive learning-rate scheduler is used for training, and gradient clipping is applied to limit exploding gradients. Figure~\ref{Fig:Bilstm} illustrates the bidirectional LSTM architecture used to estimate the local wind vector from the UAS dynamical response. The model processes a sequence of sensor-derived features and outputs the predicted north, east, and down wind components.

\begin{figure}
    \centering
    \includegraphics[width=0.75\linewidth]{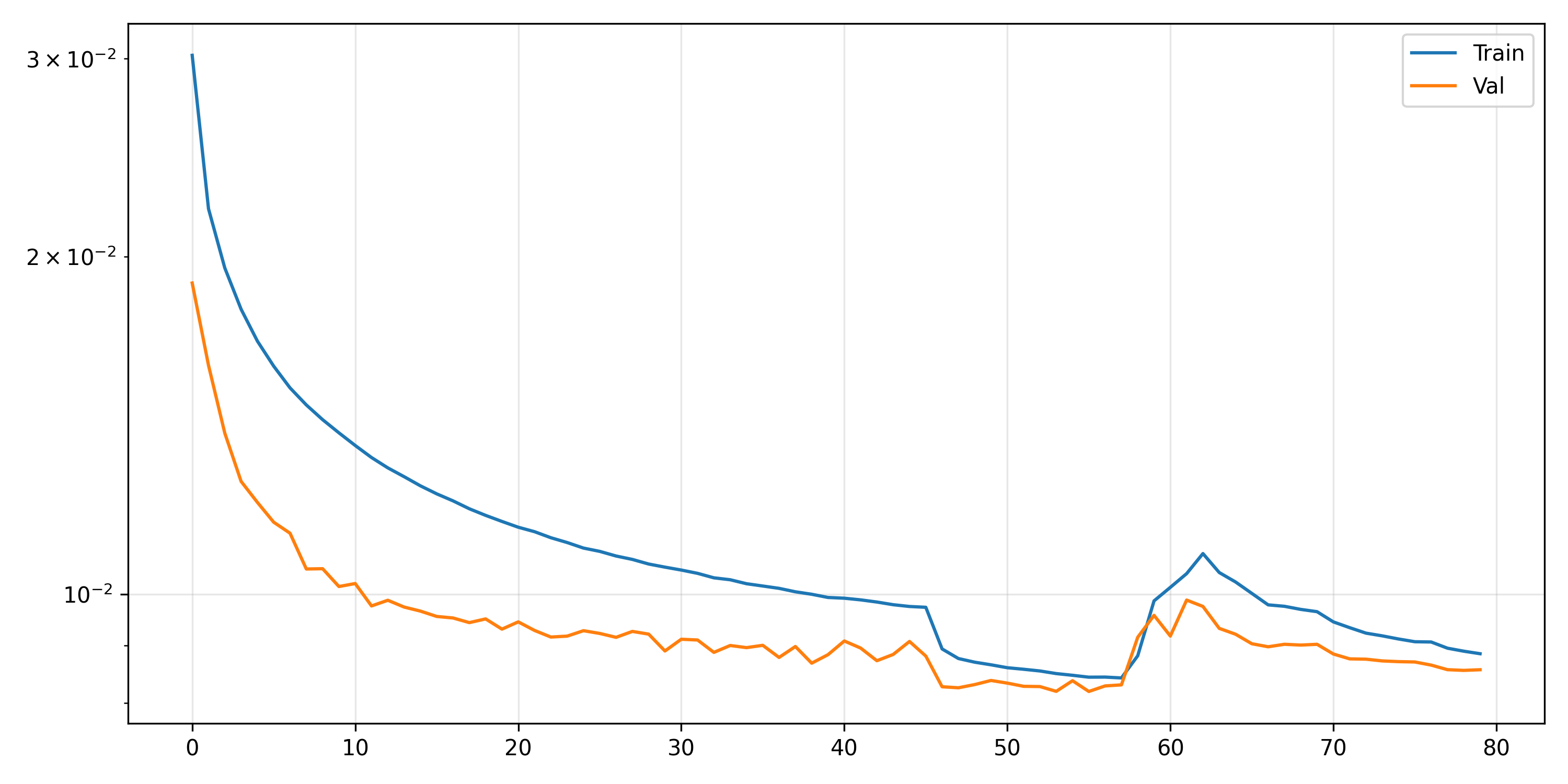}
    \caption{Training and validation MSE loss of the BiLSTM wind estimation model. 
    The monotonic decrease followed by a stable convergence trend indicates robust learning without overfitting.}
    \label{Fig:lstm_training}
\end{figure}

The network training history is shown in Figure~\ref{Fig:lstm_training}. 
The mean-squared error decreases monotonically during the initial optimization phase, followed by a mild oscillation caused by the learning-rate scheduler adjusting to flatter regions of the loss landscape. 
After this transition, both the training and validation losses continue to decrease smoothly, indicating stable convergence without signs of overfitting.

\begin{figure}
    \centering
    \includegraphics[width= 0.75\linewidth]{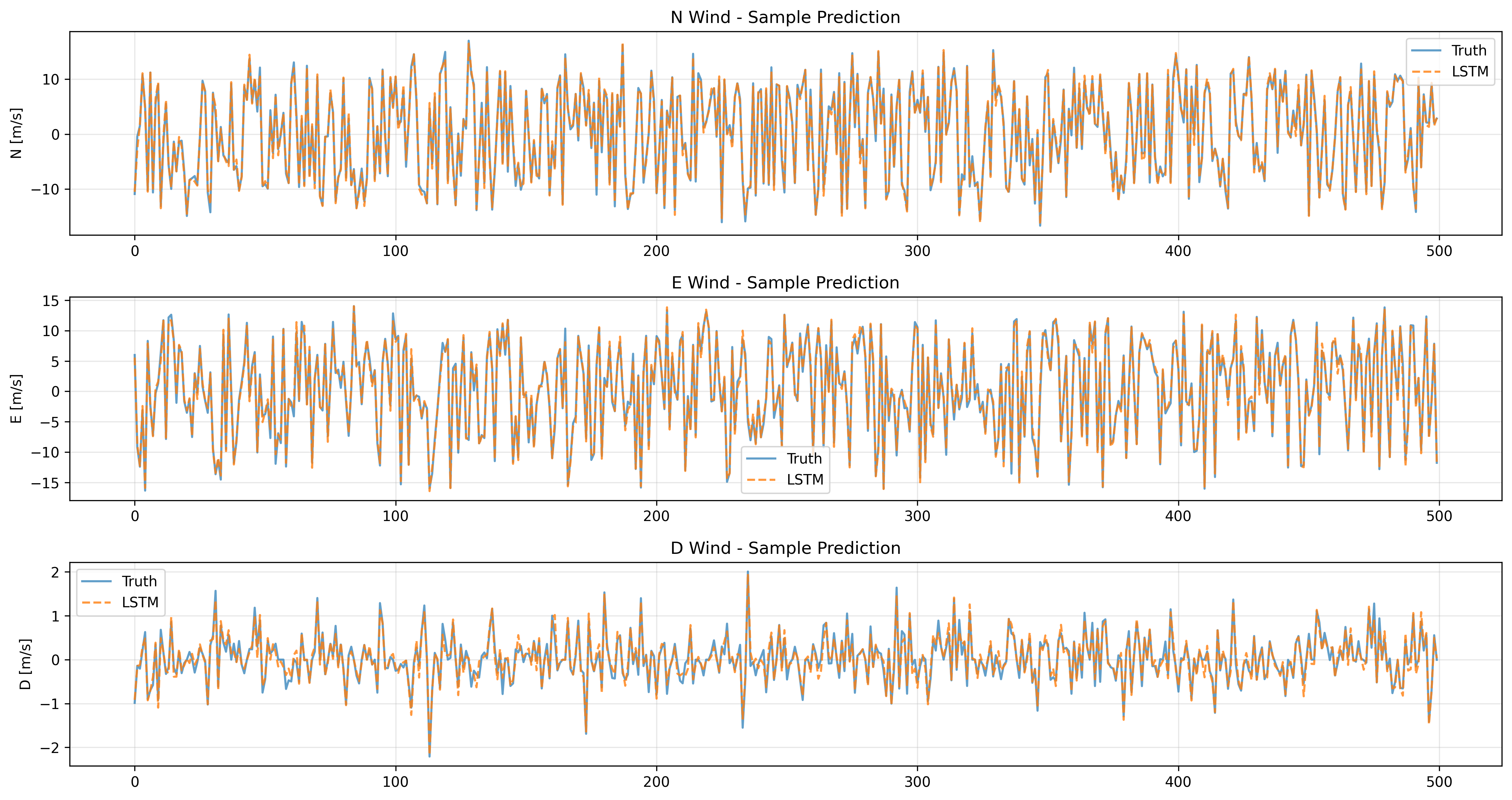}
    \caption{Example prediction of the north, east, and downwind components for the test dataset.}
    \label{Fig:lstm_sample_pred}
\end{figure}

To further illustrate the behavior of the trained model, Figure~\ref{Fig:lstm_sample_pred} presents example predictions of the north, east, and downwind components from an unseen synthetic flight. 
The LSTM prediction closely tracks the filtered truth across the full range of turbulent fluctuations, capturing both the low-frequency variations associated with mean-wind changes and the high-frequency perturbations imposed by the synthetic turbulence.

The horizontal components exhibit particularly strong agreement, while the vertical component shows slightly higher relative error. 
This behavior is expected due to the inherent difficulty of estimating vertical wind from multirotor dynamics at flux scales; however, the magnitude of this error remains within the convective-scale tolerance commonly accepted in the meteorological community.

\subsection{Local Wind Estimation Performance}

After training, the model is deployed on previously unseen synthetic flights and evaluated independently for each UAS. 
During inference, sliding windows of normalized features are passed through the network to obtain predicted wind components, 
\begin{equation}
    \hat{\mathbf{w}}(t) = 
    f_{\theta}\!\left(\tilde{\mathbf{x}}(t-L+1:t)\right),
\end{equation}
followed by inverse normalization to recover the physical wind vector.
Performance is quantified using the component-wise root-mean-square error (RMSE),
\begin{equation}
    \mathrm{RMSE}_k = 
    \sqrt{
        \frac{1}{N}
        \sum_{i=1}^{N}
        \left(\hat{U}_k^{(i)} - U_k^{(i)}\right)^2
    },
    \qquad k \in \{N,E,D\},
\end{equation}
and the relative error,
\begin{equation}
    \mathrm{RE}_k =
    \frac{
        \frac{1}{N}\sum_{i=1}^{N}
        \left|\hat{U}_k^{(i)} - U_k^{(i)}\right|
    }{
        \frac{1}{N}\sum_{i=1}^{N}\left|U_k^{(i)}\right|
    }\times 100~\%.
\end{equation}

To assess the robustness of the estimator across the range of atmospheric conditions encountered in the synthetic environment, 
three representative wind regimes are examined: \textit{low-wind}, \textit{moderate-wind}, and \textit{high-wind} cases. 
For each regime, RMSE and relative error are computed separately for the north, east, and down components. 
The horizontal components (\(U_N\) and \(U_E\)) consistently exhibit the lowest error, reflecting the strong excitation of horizontal dynamics and the clear coupling between horizontal wind and UAS motion. 
In contrast, the vertical component (\(U_D\)) displays slightly larger relative error due to its smaller signal magnitude and weaker dynamical observability in multirotor platforms. 
However, the magnitude of the vertical-wind error remains within the convective-scale tolerance commonly accepted in the meteorological community, 
and is reported directly in units of m\,s\(^{-1}\), as UAS observations at this scale are intended for boundary-layer characterization rather than flux measurements.

\begin{figure}
    \centering
    \includegraphics[width=0.95\linewidth]{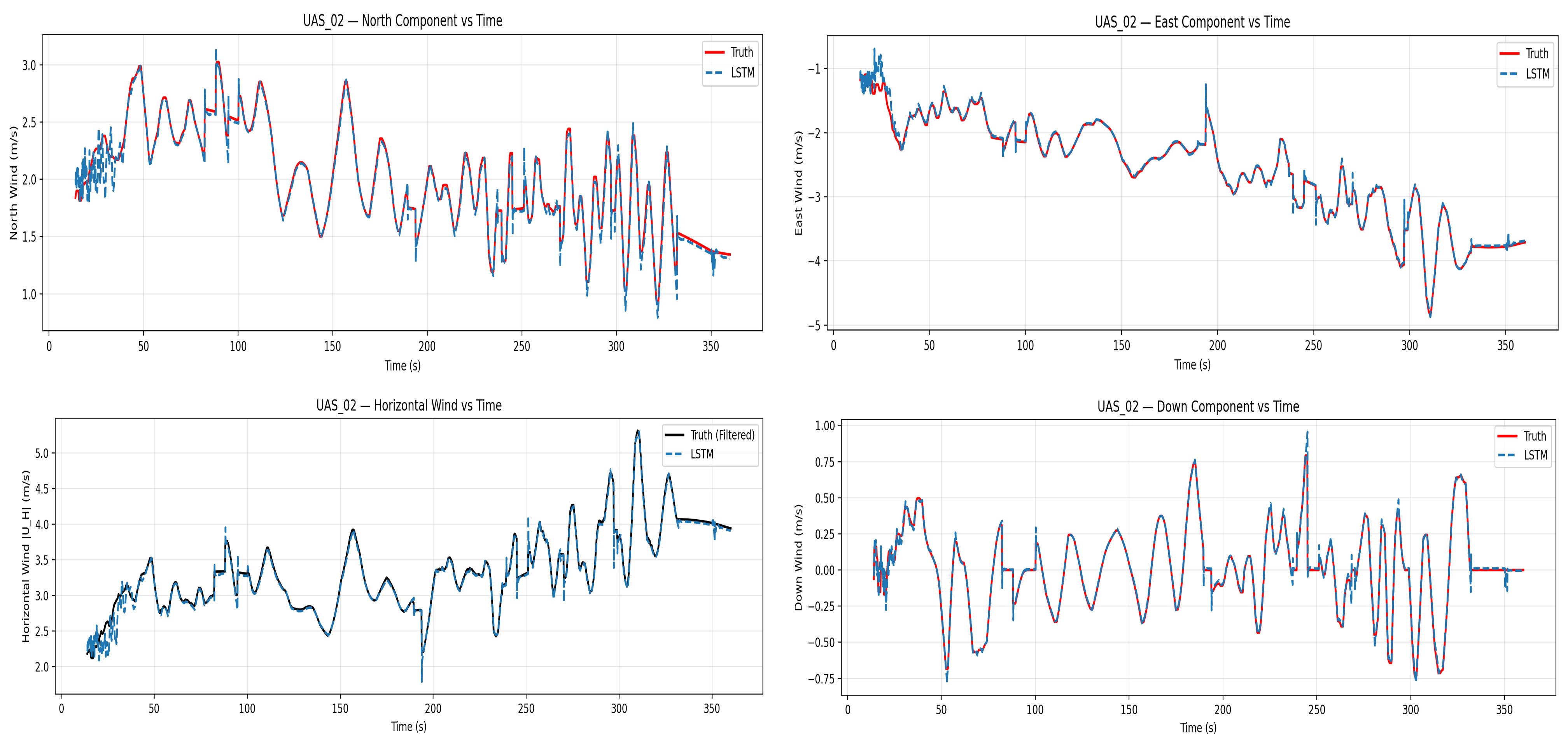}
    \caption{
        Predicted and true wind components 
        $\left[U_N,\,U_E,\,U_D\right]$ and the horizontal wind for the low-wind regime. 
        The estimator captures both the mean-flow structure and the small-scale variability. 
        Reported performance metrics:
        RMSE $= \{0.064,\,0.062,\,0.029\}\,\text{m\,s}^{-1}$,
        and relative errors $= \{1.6,\,1.2,\,6.1\}\,\%$.
    }
    \label{Fig:lstm_low_wind}
\end{figure}

\begin{figure}
    \centering
    \includegraphics[width=0.95\linewidth]{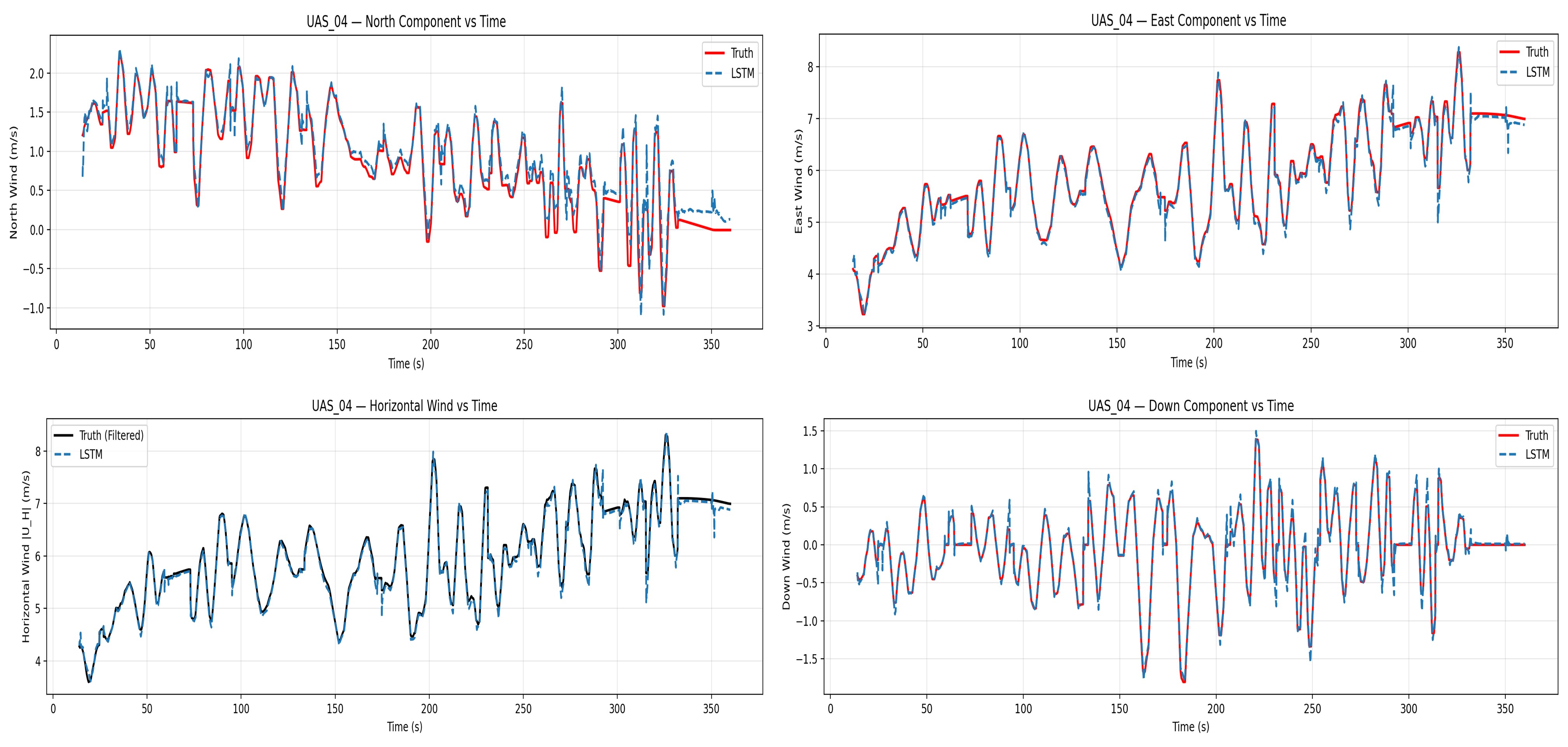}
    \caption{Predicted and true wind components 
        $\left[U_N,\,U_E,\,U_D\right]$ and the horizontal wind for the moderate-wind regime. 
        The estimator captures both the mean-flow structure and the small-scale variability. 
        Reported performance metrics:
        RMSE $= \{0.122,\,0.129,\,0.061\}\,\text{m\,s}^{-1}$,
        and relative errors $= \{9.5,\,1.4,\,8.2\}\,\%$.
    }
    \label{Fig:lstm_mod_wind}
\end{figure}

\begin{figure}
    \centering
    \includegraphics[width=0.95\linewidth]{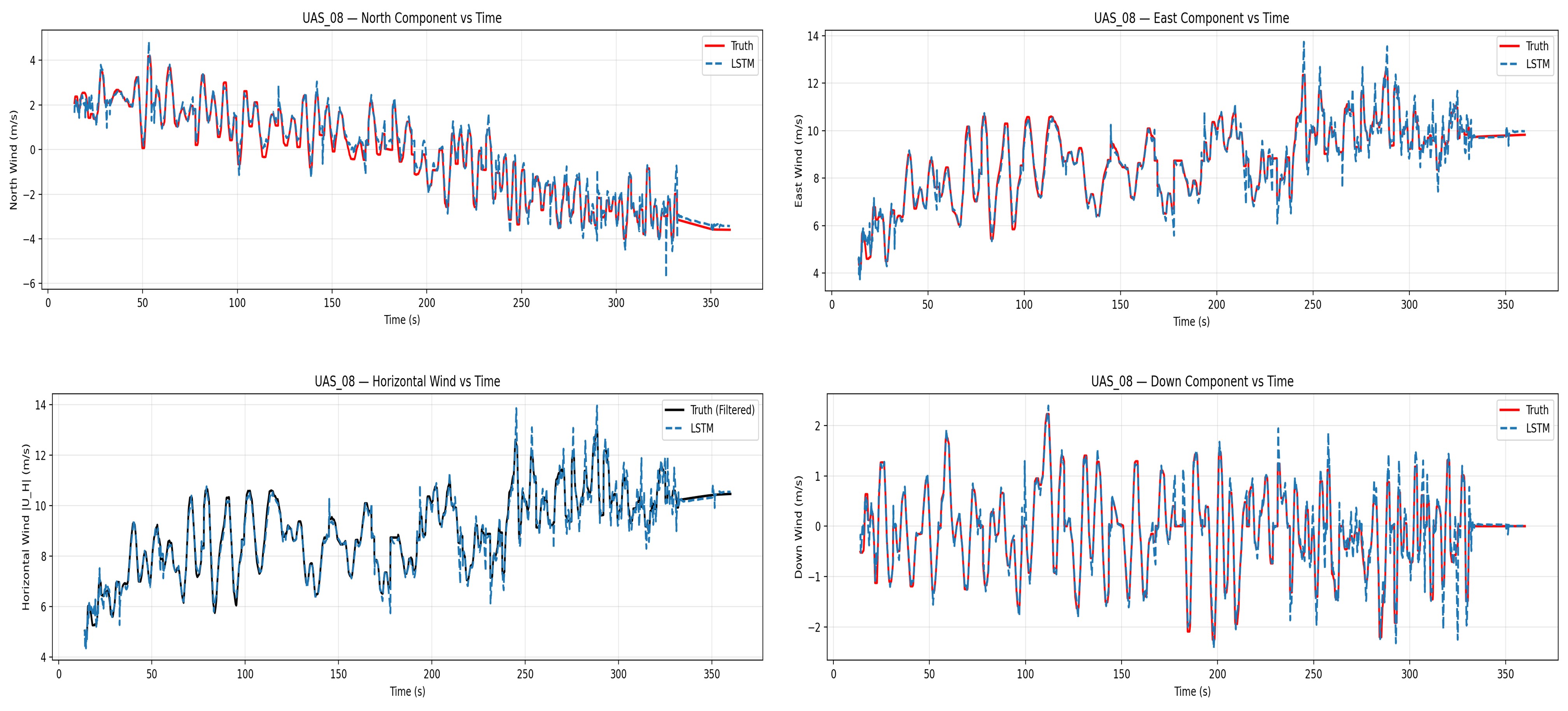}
    \caption{
    Predicted and true wind components 
        $\left[U_N,\,U_E,\,U_D\right]$ and the horizontal wind for the high-wind regime. 
        The estimator captures both the mean-flow structure and the small-scale variability. 
        Reported performance metrics:
        RMSE $= \{0.273,\,0.271,\,0.091\}\,\text{m\,s}^{-1}$,
        and relative errors $= \{11.1,\,2.0,\,15.3\}\,\%$.
    }
    \label{Fig:lstm_high_wind}
    
\end{figure}

To examine the performance of the local wind estimator under different flow conditions, 
three representative wind regimes are selected from the synthetic dataset: 
a low-wind case Figure \ref{Fig:lstm_low_wind}, a moderate-wind case Figure \ref{Fig:lstm_mod_wind}, and a high-wind Figure \ref{Fig:lstm_high_wind} case. 
For each regime, the predicted and filtered wind components 
$\left[U_N,\,U_E,\,U_D\right]$ are compared, and the corresponding RMSE 
and relative-error values are reported in the figure captions. 
These examples illustrate how the BiLSTM model responds to increasing mean-wind 
and turbulence intensity, highlighting its ability to reproduce horizontal wind 
components with high accuracy while retaining physically meaningful performance 
in the vertical component, which is inherently more challenging for multirotor platforms 
operating at convective scales.


In summary, the BiLSTM model provides accurate and stable estimates of the horizontal wind components across all examined wind regimes. 
Both the northward and eastward winds are consistently predicted with low RMSE and relative error under low-, moderate-, and high-wind conditions, demonstrating the robustness of the learning-based approach for horizontal wind estimation using UAS dynamics.

The estimation accuracy of the vertical wind component is comparatively lower, which is expected for two primary reasons. 
First, vertical wind variability is traditionally characterized at flux scales, whereas multirotor UAS platforms are designed to operate at convective scales, making direct inference of fine-scale vertical motions inherently challenging. 
Second, the simulated vertical wind is dominated by turbulence-induced fluctuations with strong stochastic behavior, which are difficult to reconstruct reliably from vehicle motion alone. 
Nevertheless, the resulting vertical-wind errors remain within acceptable bounds for convective-scale atmospheric applications.

\section{Physics-Informed Neural Network for Wind-Field Reconstruction}

\subsection{Problem Formulation and Objectives}
\label{sec:pinn_formulation}

While the data-driven model described in section \ref{lstm} provides estimates of the local wind components along individual UAS trajectories, many applications in atmospheric profiling and boundary-layer studies require a continuous representation of the wind field in both space and time. This motivates the formulation of a wind-field reconstruction problem, in which sparse, trajectory-based wind estimates are interpolated into a four-dimensional (4D) wind field defined over a spatial domain and a finite time window.

The reconstruction problem addressed in this work is inherently inverse and underdetermined. Wind observations are available only along the flight paths of a limited number of UAS platforms, and no direct measurements exist throughout the majority of the domain. Moreover, the atmospheric flow is unsteady and exhibits turbulent fluctuations, which cannot be uniquely determined from sparse measurements alone. As a result, classical interpolation or regression approaches are insufficient, as they either fail to enforce physical consistency or excessively smooth the reconstructed flow.

To address these challenges, a physics-informed neural network (PINN) framework is employed to reconstruct a continuous wind field from sparse local wind estimates. The objective of the PINN is not to solve the full Navier--Stokes equations, but rather to infer a physically plausible wind field that (i) remains consistent with the data-driven wind estimates along UAS trajectories, (ii) satisfies weak physical constraints representative of atmospheric flow, and (iii) preserves unsteady and turbulent flow features supported by the available data. This formulation enables the reconstruction of a spatially and temporally continuous wind field while avoiding the introduction of artificial boundary conditions or overly restrictive physical assumptions.

\subsection{Network Inputs and Outputs}
\label{sec:pinn_io}

The PINN is formulated as a regression model that maps spatiotemporal coordinates to the three-dimensional wind velocity components. The network input consists of the four-dimensional vector
\begin{equation}
\boldsymbol{x} = \left( x, y, z, t \right),
\end{equation}
where \(x\) and \(y\) denote the horizontal spatial coordinates, \(z\) represents altitude above ground level, and \(t\) denotes time. These coordinates are expressed in a common Earth-fixed reference frame consistent with the UAS trajectory data.

The network output corresponds to the reconstructed wind velocity vector
\begin{equation}
\boldsymbol{u}(\boldsymbol{x}) = \left( U_N, U_E, U_D \right),
\end{equation}
where \(U_N\), \(U_E\), and \(U_D\) denote the northward, eastward, and downward wind components, respectively. The target values used during training are obtained from the data-driven local wind estimates produced by the BiLSTM model described in Section \ref{lstm}. These estimates retain unsteady and turbulent fluctuations present in the simulated atmospheric flow and serve as sparse observations for the reconstruction process.

By learning a continuous mapping from \((x,y,z,t)\) to \((U_N, U_E, U_D)\), the PINN yields a unified wind-field representation that can be queried at arbitrary locations and times within the reconstruction domain, enabling the generation of spatial slices, vertical profiles, and time-resolved wind fields.

\subsection{Feature Representation and Normalization}
\label{sec:pinn_features}

Prior to training, the input coordinates \((x,y,z,t)\) are normalized to improve numerical stability and convergence of the neural network. The spatial coordinates \((x,y,z)\) are standardized using their respective mean and standard deviation computed over the reconstruction domain, while time is independently scaled to the interval \([-1,1]\) to avoid temporal and spatial scales during training.

To enhance the the model's ability to represent unsteady flow features and vertical variability, Fourier feature embeddings are applied to the temporal and vertical coordinates. Specifically, sinusoidal basis functions of the form
\begin{equation}
\sin\left(2\pi k \frac{t}{T}\right), \quad \cos\left(2\pi k \frac{t}{T}\right),
\end{equation}
and
\begin{equation}
\sin\left(2\pi k \frac{z}{Z}\right), \quad \cos\left(2\pi k \frac{z}{Z}\right),
\end{equation}
are appended to the normalized input vector, where \(T\) and \(Z\) denote the temporal and vertical extents of the reconstruction domain, and \(k\) represents the selected Fourier modes.

Fourier features are introduced only in the time and vertical dimensions, as these directions are most strongly associated with atmospheric unsteadiness and vertical shear in the boundary layer. Horizontal coordinates are retained in their normalized form to avoid introducing artificial periodicity in the lateral directions. This feature representation allows the PINN to capture multi-scale temporal variability and height-dependent flow structure while maintaining a compact and physically interpretable input space.

\subsection{Loss Function and Physical Regularization}
\label{sec:pinn_loss}

The PINN is trained by minimizing a composite loss function that combines data fidelity with weak physics regularization. The objective is to reconstruct a wind field that remains consistent with the sparse local wind estimates while discouraging nonphysics behavior, without imposing overly restrictive constraints that would suppress turbulence.

The primary component of the loss function is a data-misfit term,
\begin{equation}
\mathcal{L}_{\mathrm{data}} =
\frac{1}{N}
\sum_{i=1}^{N}
\left\|
\boldsymbol{u}_{\mathrm{PINN}}(\boldsymbol{x}_i)
-
\boldsymbol{u}_{\mathrm{est}}(\boldsymbol{x}_i)
\right\|^2,
\end{equation}
where \(\boldsymbol{u}_{\mathrm{est}} = (U_N, U_E, U_D)\) denotes the local wind estimates obtained from the LSTM model at the UAS measurement locations \(\boldsymbol{x}_i = (x_i,y_i,z_i,t_i)\), and \(N\) is the number of available data points. This term anchors the reconstructed wind field to the data-driven estimates along the UAS trajectories.

To promote physics consistency, a weak divergence-free constraint is applied throughout the reconstruction domain. The corresponding loss term is defined as
\begin{equation}
\mathcal{L}_{\mathrm{div}} =
\frac{1}{M}
\sum_{j=1}^{M}
\left(
\frac{\partial U_N}{\partial x}
+
\frac{\partial U_E}{\partial y}
+
\frac{\partial U_D}{\partial z}
\right)^2
\Bigg|_{\boldsymbol{x}_j},
\end{equation}
where the derivatives are evaluated at randomly sampled collocation points \(\boldsymbol{x}_j\) within the domain using differentiation. This term reflects the approximately incompressible nature of atmospheric flow at the scales of interest, while remaining sufficiently weak to avoid suppressing turbulent fluctuations.

In addition, a near-ground smoothness penalty is introduced to discourage nonphysical vertical gradients in the lowest portion of the domain. This term penalizes large values of \(\partial U_D / \partial z\) below a prescribed height threshold and serves as a soft regularization near the surface without enforcing explicit boundary conditions.

The total loss minimized during training is given by
\begin{equation}
\mathcal{L}
=
\mathcal{L}_{\mathrm{data}}
+
\lambda_{\mathrm{phys}} \, \mathcal{L}_{\mathrm{div}}
+
\lambda_{\mathrm{smooth}} \, \mathcal{L}_{\mathrm{smooth}},
\end{equation}
where \(\lambda_{\mathrm{phys}}\) and \(\lambda_{\mathrm{smooth}}\) are weighting coefficients controlling the influence of the physical regularization terms. These weights are selected to ensure that the data term remains dominant, allowing the reconstructed wind field to retain unsteady and turbulent features supported by the observations, while preventing the emergence of clearly nonphysical solutions.

\subsection{Training Strategy and Reconstruction Workflow}
\label{sec:pinn_training}

The PINN is trained using a supervised learning strategy in which the data-driven local wind estimates serve as sparse interior observations, while physical regularization terms guide the reconstruction toward physically plausible solutions. To ensure sufficient representational capacity for reconstructing complex, unsteady flow structures, the network is intentionally over-parameterized relative to the amount of available data. This design choice allows the model to interpolate sparse observations without imposing excessive smoothness on the reconstructed wind field.

During training, the data-misfit term is maintained as the dominant component of the loss function, while the physical regularization terms are applied with relatively small weighting coefficients. This weak-physics formulation reflects the objective of the reconstruction task: rather than enforcing a full set of governing equations, the PINN is guided to satisfy minimal physical consistency while retaining turbulent features present in the data. The divergence-free constraint acts as a global regularizer that suppresses clearly nonphysical behavior, whereas the near-ground smoothness penalty stabilizes the solution close to the surface without constraining the flow elsewhere in the domain.

Physical constraints are evaluated at randomly sampled collocation points throughout the spatiotemporal domain, ensuring that the regularization is distributed across the interior rather than concentrated near data locations. This sampling strategy promotes global consistency of the reconstructed field while avoiding bias toward any specific region of the domain.

Once trained, the PINN provides a continuous four-dimensional wind-field representation that can be queried at arbitrary spatial locations and times within the reconstruction domain. The reconstructed wind field is subsequently analyzed through horizontal and vertical slices, vertical profiles, and time-resolved comparisons with the simulated reference data. This workflow enables both qualitative visualization of the reconstructed flow structure and quantitative assessment of reconstruction accuracy, forming the basis for the results presented in the following section.

\section{Results}
\label{sec:results}

\subsection{Wind-Field Reconstruction Under Moderate Wind Conditions}
\label{sec:results_moderate}

This subsection presents a representative wind-field reconstruction obtained under moderate wind conditions using measurements from a formation of nine UAS platforms. The purpose of this case is to illustrate the reconstruction process and the resulting wind-field structure produced by the proposed framework, prior to quantitative comparisons across different UAS configurations.

The nine UAS platforms collectively sample the reconstruction domain through coordinated ascent trajectories, providing sparse, time-resolved interior observations of the local wind field. These data-driven wind estimates are assimilated by the PINN to infer a continuous four-dimensional wind field over the spatial domain and time window of interest. No wind information is prescribed outside the UAS trajectories, requiring the PINN to interpolate the flow field based solely on the available measurements and weak physical regularization.

The reconstructed wind field is evaluated qualitatively against the simulated reference wind generated using the von Kármán turbulence model. Horizontal and vertical slices are used to examine the spatial organization of the flow, vertical shear, and unsteady features supported by the observations. This case serves as a baseline demonstration of the reconstruction capability using a moderate number of UAS platforms, while more detailed quantitative comparisons across different platform configurations are presented in the following subsection.

\begin{figure}[t]
    \centering
    \includegraphics[width=0.9\linewidth]{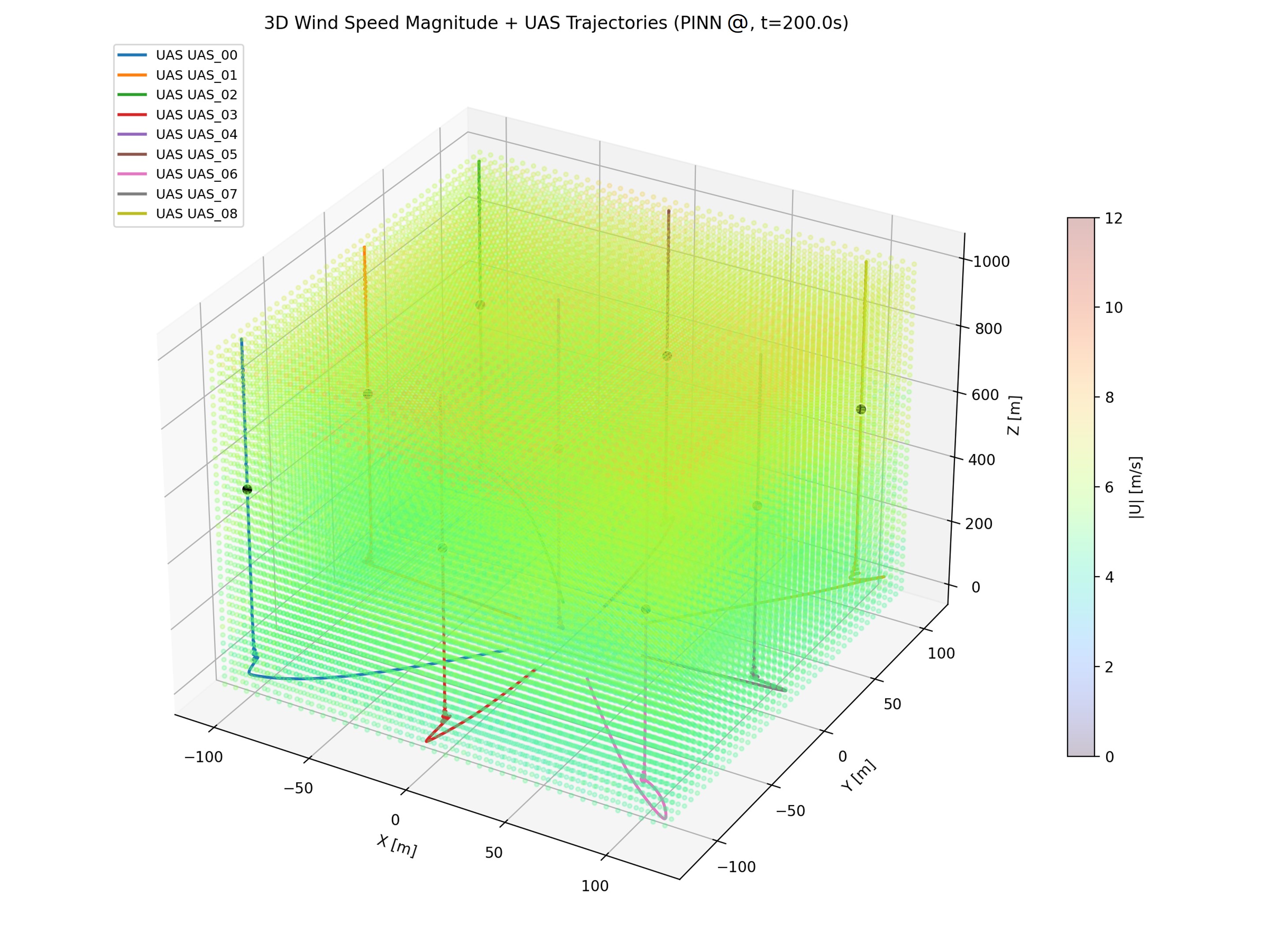}
    \caption{Three-dimensional reconstruction of wind-speed magnitude at \(t = 200~\mathrm{s}\) using nine UAS platforms. Colored points represent the reconstructed wind magnitude, while solid lines indicate the UAS ascent trajectories.}
    \label{fig:pinn_3d_moderate}
\end{figure}

Figure~\ref{fig:pinn_3d_moderate} presents a three-dimensional visualization of the reconstructed wind-speed magnitude at \(t = 200~\mathrm{s}\), together with the ascent trajectories of the nine UAS platforms. Although wind observations are available only along these sparse trajectories, the PINN reconstructs a spatially continuous wind field throughout the entire domain. The reconstructed volume exhibits coherent vertical organization and gradual lateral variation, indicating that the model effectively interpolates interior measurements into a physically plausible volumetric flow representation.

\begin{figure}
    \centering
    \includegraphics[width=1.0\linewidth]{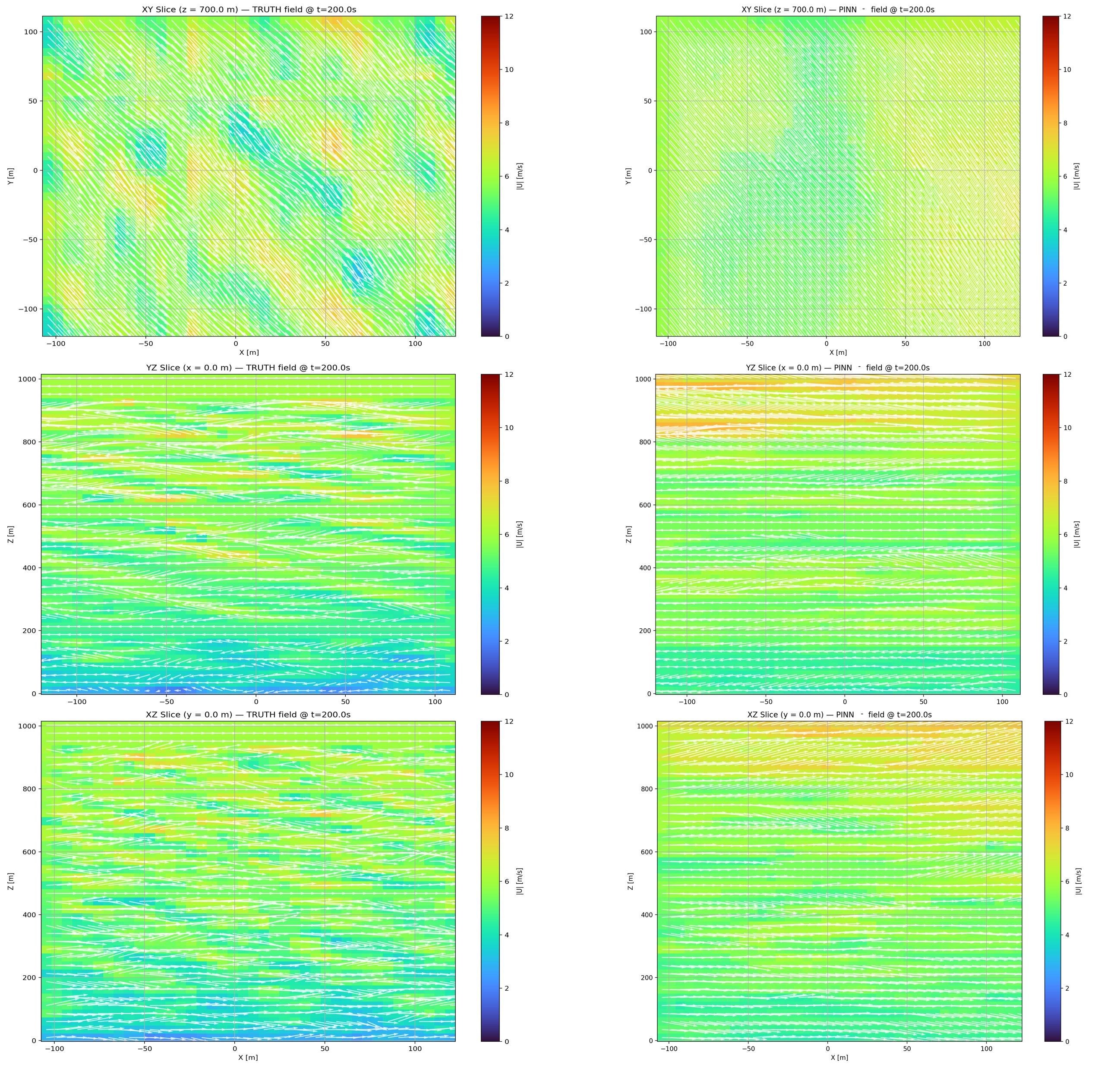}
    \caption{Comparison of reference and reconstructed wind fields at \(t = 200~\mathrm{s}\). Left: reference wind field generated using the von Kármán turbulence model. Right: PINN-reconstructed wind field using nine UAS platforms. Rows correspond to \(x\text{--}y\) (\(z = 700~\mathrm{m}\)), \(y\text{--}z\) (\(x = 0~\mathrm{m}\)), and \(x\text{--}z\) (\(y = 0~\mathrm{m}\)) slices. The RMSE at UAS locations for north-, east-, and down-wind components are 0.105, 0.100, and 0.076~m~s\(^{-1}\), respectively, with an overall RMSE of 1.11~m~s\(^{-1}\).}
    \label{fig:pinn_slices_moderate}
\end{figure}

To further evaluate the reconstructed wind field, horizontal and vertical slices are compared against the simulated reference solution generated using the von Kármán turbulence model. Figure~\ref{fig:pinn_slices_moderate} shows representative \(XY\), \(YZ\), and \(ZX\) slices extracted at the same time instant for both the reference and reconstructed fields. The horizontal slice at \(z = 700~\mathrm{m}\) demonstrates that the PINN captures the dominant flow direction and large-scale spatial gradients present in the reference field. While the reference solution exhibits pronounced small-scale variability associated with turbulent structures, these high-frequency fluctuations are partially attenuated in the reconstruction.

Vertical slices in the \(x\text{--}z\) and \(y\text{--}z\) planes further illustrate the reconstruction performance. The PINN successfully recovers the height-dependent variation and vertical shear structure of the wind field across the domain. Compared to the reference solution, the reconstructed field exhibits smoother transitions in both the vertical and lateral directions, reflecting the influence of sparse sampling and weak physical regularization. Importantly, the reconstructed field maintains consistent vertical organization, indicating that information from multiple UAS trajectories is effectively integrated to infer the three-dimensional flow structure.

\begin{figure}
    \centering
    \includegraphics[width=0.9\linewidth]{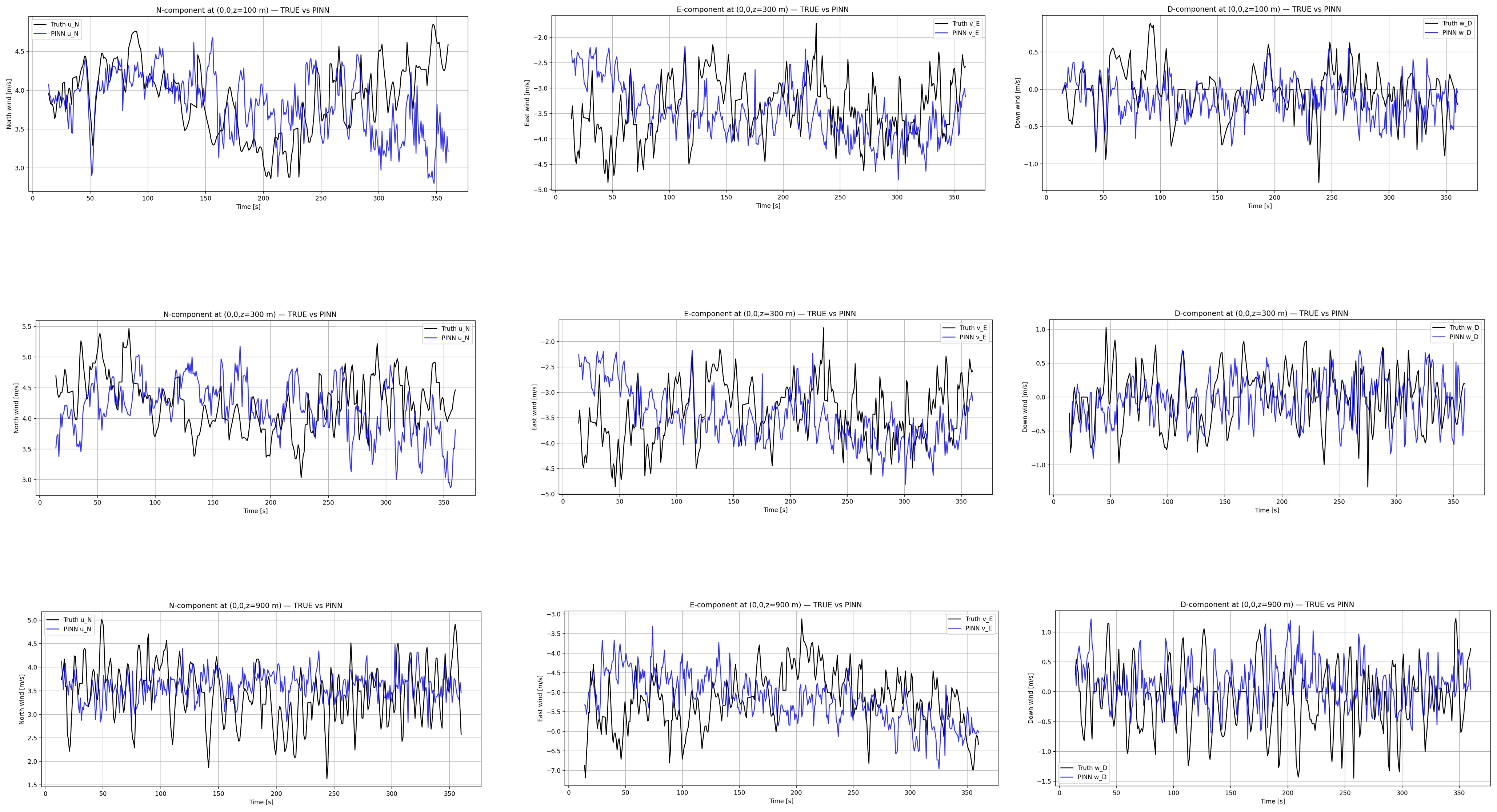}
    \caption{Time-series comparison of reconstructed and reference wind components at fixed spatial locations and selected heights. The PINN reconstruction captures the mean behavior and dominant unsteady trends while exhibiting reduced high-frequency variability relative to the reference field.}
    \label{fig:pinn_timeseries_moderate}
\end{figure}

Time-series comparisons of the reconstructed and reference wind components at fixed spatial locations and selected heights are shown in Figure~\ref{fig:pinn_timeseries_moderate}. The PINN reconstruction closely follows the low-frequency temporal evolution of the reference wind components, capturing the mean behavior and dominant unsteady trends in the north, east, and vertical directions. High-frequency fluctuations present in the reference signals are reduced in amplitude in the reconstructed time series, particularly for the vertical wind component, which is the least constrained by the available observations.

This smoothing behavior is consistent with the reconstruction objective and the weak physical regularization applied during training. The PINN favors coherent flow structures supported by the sparse measurements while suppressing unresolved turbulent fluctuations that cannot be reliably inferred from the available data.

Overall, the nine-UAS reconstruction demonstrates that the proposed PINN framework can recover the dominant spatial and temporal structure of a moderate wind field using sparse interior observations alone. The reconstructed wind field preserves mean flow direction, vertical shear, and large-scale variability while exhibiting controlled smoothing of small-scale turbulence. This case serves as a baseline demonstration of the reconstruction capability, with quantitative accuracy and sensitivity to UAS configuration examined in the following subsection.

\subsection{Effect of UAS Configuration on Reconstruction Accuracy}
\label{sec:uas_config_rmse}

An important practical objective of the proposed wind-field reconstruction framework is to identify a UAS configuration that achieves sufficient reconstruction accuracy while minimizing deployment cost and operational complexity. In this context, both the number of UAS platforms and their flight trajectories directly influence the spatial and temporal coverage of the reconstruction domain.

To evaluate this trade-off, several UAS configurations with varying platform counts were examined, and the corresponding wind-field reconstructions were performed using the same training and inference procedure described in Section~\ref{sec:pinn_training}. For each configuration, the reconstructed wind field was compared against the simulated reference wind, and the RMSE was computed over the reconstruction domain.

Table~\ref{tab:uas_config_rmse} summarizes the UAS configurations which are considered in this study and their respective reconstruction errors. The configurations range from sparse deployments with four UAS platforms to denser formations consisting of up to sixteen platforms. These results provide a quantitative basis for assessing the cost-effectiveness of different UAS deployments and inform the selection of an appropriate configuration for practical wind-field reconstruction.

\begin{table}[t]
\centering
\caption{Reconstruction RMSE for different UAS configurations under moderate wind conditions.}
\label{tab:uas_config_rmse}
\begin{tabular}{c c c c c}
\hline
\textbf{UAS Configuration} & \textbf{RMSE$_N$} & \textbf{RMSE$_E$} & \textbf{RMSE$_D$} & \textbf{RMSE$_{overall}$} \\
 & (m s$^{-1}$) & (m s$^{-1}$) & (m s$^{-1}$) & (m s$^{-1}$) \\
\hline
4 UAS  & 0.111 & 0.117 & 0.078 & 0.125 \\
5 UAS  & 0.114 & 0.092 & 0.064 & 0.118 \\
6 UAS  & 0.149 & 0.095 & 0.098 & 0.144 \\
7 UAS  & 0.099 & 0.109 & 0.070 & 0.128 \\
9 UAS  & 0.105 & 0.100 & 0.076 & 0.151 \\
12 UAS & 0.149 & 0.151 & 0.095 & 0.154 \\
\hline
\end{tabular}
\end{table}

The results in Table~\ref{tab:uas_config_rmse} indicate that reconstruction accuracy does not improve monotonically with increasing number of UAS platforms. Instead, the error exhibits a non-linear dependence on platform count, reflecting the interaction between sampling density, data consistency, and the weakly constrained nature of the inverse reconstruction problem.

Among the configurations considered, the five-UAS case yields the lowest overall RMSE, while the seven-UAS configuration provides the best performance for the northward wind component. These results suggest that moderate UAS counts can provide sufficient spatial and temporal coverage to constrain the dominant wind-field structure without introducing excessive redundancy or conflicting information.

As the number of UAS platforms increases beyond this range, the overall reconstruction error increases. This behavior can be attributed to several factors. First, additional UAS trajectories introduce measurements at different locations and times that may sample distinct turbulent realizations within the flow. Because small-scale turbulence is weakly constrained in the PINN formulation, inconsistencies among densely sampled trajectories can increase the difficulty of fitting a single smooth, physically regularized wind field.

Second, higher UAS counts effectively increase the density of data constraints without a corresponding increase in independent physical constraints. In a weakly regularized inverse problem, this can lead to over-conditioning, where the PINN must reconcile mutually incompatible local estimates, resulting in increased global error. This effect is particularly pronounced for the vertical wind component, which is the least directly observed and most sensitive to noise and trajectory-dependent estimation errors.

These results highlight an important practical consideration: beyond a certain platform count, additional UAS deployments do not necessarily improve reconstruction accuracy and may instead degrade performance. Consequently, an intermediate UAS configuration represents a more cost-effective and robust solution for wind-field reconstruction under moderate wind conditions.

\section{Conclusions and Future Work}

This work presented a framework for reconstructing 4D atmospheric wind fields using measurements obtained exclusively from a coordinated swarm of UAS. The approach combines a synthetic turbulence environment, high-fidelity multirotor UAS simulation, data-driven local wind estimation from UAS vehicle dynamics, and a PINN model to reconstruct a continuous wind field in space and time without relying on dedicated wind sensors or fixed sensing infrastructure.

Local wind components in the north, east, and down directions were estimated using a bidirectional LSTM model trained on synthetic flights spanning a range of atmospheric boundary layer conditions. The estimator demonstrated robust performance across low, moderate, and high wind regimes, with consistently strong accuracy in the horizontal components. These local wind estimates served as sparse interior observations for the physics-informed reconstruction.

The PINN successfully reconstructed the dominant spatial and temporal structure of the mean wind field up to 1000 m altitude. The reconstructed fields preserved mean flow direction, vertical shear, and large-scale unsteady behavior, while exhibiting controlled smoothing of unresolved turbulent fluctuations. An analysis of UAS configuration showed that reconstruction accuracy does not improve monotonically with increasing platform count, and that intermediate swarm sizes can provide a more cost-effective and robust solution.

Future work will focus on extending the framework to more complex atmospheric conditions and improving sensitivity to vertical wind components. Incorporating alternative turbulence models, refining local wind estimation using additional onboard features, and exploring adaptive multi-UAS trajectory design are promising directions. Validation using real-world flight data will be essential to assess performance under operational conditions and to support deployment in practical atmospheric sensing applications.

\bibliographystyle{elsarticle-num}
\bibliography{references}

\end{document}